\pdfoutput=1

\documentclass[11pt]{article}

\usepackage[final]{acl}

\usepackage{times}
\usepackage{latexsym}

\usepackage[T1]{fontenc}

\usepackage[utf8]{inputenc}

\usepackage{microtype}

\usepackage{inconsolata}

\usepackage{amsmath,amsfonts,bm}

\def\eqref#1{equation~\ref{#1}}

\def\1{\bm{1}}

\def\vs{{\bm{s}}}

\def\vx{{\bm{x}}}

\DeclareMathAlphabet{\mathsfit}{\encodingdefault}{\sfdefault}{m}{sl}
\SetMathAlphabet{\mathsfit}{bold}{\encodingdefault}{\sfdefault}{bx}{n}

\usepackage{duckuments}
\usepackage{graphicx}
\usepackage{dsfont}

\usepackage{booktabs}
\usepackage[para,online,flushleft]{threeparttable}
\usepackage{multicol}
\usepackage{multirow}
\usepackage{caption}
\usepackage{subcaption}
\usepackage{adjustbox}
\usepackage{colortbl}

\usepackage{wrapfig}

\usepackage{makecell}

\usepackage[skip=6pt]{caption} 
\usepackage{titlesec}

\usepackage[capitalize]{cleveref}

\Crefname{figure}{{Figure}}{{Figures}}

\title{Shifting Attention to Relevance: Towards the Predictive Uncertainty Quantification of Free-Form Large Language Models}

\author{
  Jinhao Duan$^1$ ~~~~ Hao Cheng$^3$ ~~~~ Shiqi Wang$^2$ ~~~~ Alex Zavalny$^1$ ~~~~ Chenan Wang$^1$ \\
  {\bf Renjing Xu$^3$ ~~~~ Bhavya Kailkhura$^4$ ~~~~ Kaidi Xu$^1$\thanks{~~Corresponding author: Kaidi Xu <kx46@drexel.edu>.}} \\
   \\
  $^1$Drexel University 
  $^2$AWS AI Lab \\
  $^3$Hong Kong University of Science and Technology (Guangzhou)\\
  $^4$Lawrence Livermore National Laboratory\\
}

\definecolor{ao(english)}{rgb}{0.0, 0.5, 0.0}
\definecolor{amaranth}{rgb}{0.9, 0.17, 0.31}

\newcolumntype{?}{!{\vrule width 1pt}}

\begin{document}
\maketitle

\begin{abstract}

Large Language Models (LLMs) show promising results in language generation and instruction following but frequently ``hallucinate'', making their outputs less reliable. Despite Uncertainty Quantification's (UQ) potential solutions, implementing it accurately within LLMs is challenging. Our research introduces a simple heuristic: not all tokens in auto-regressive LLM text equally represent the underlying meaning, as ``linguistic redundancy'' often allows a few keywords to convey the essence of long sentences. However, current methods underestimate this inequality when assessing uncertainty, causing tokens with limited semantics to be equally or excessively weighted in UQ. To correct this, we propose \textbf{S}hifting \textbf{A}ttention to more \textbf{R}elevant (\textit{SAR}) components at both token- and sentence-levels for better UQ. We conduct extensive experiments involving a range of popular ``off-the-shelf” LLMs, such as Vicuna, WizardLM, and LLaMA-2-chat, with model sizes extending up to 33B parameters. We evaluate various free-form question-answering tasks, encompassing domains such as reading comprehension, science Q\&A, and medical Q\&A. Our experimental results, coupled with a comprehensive demographic analysis, demonstrate the superior performance of \textit{SAR}. The code is available at \url{https://github.com/jinhaoduan/SAR}.

\end{abstract}

\section{Introduction}

\begin{figure}[t]
    \centering
    \includegraphics[width=0.9\linewidth]{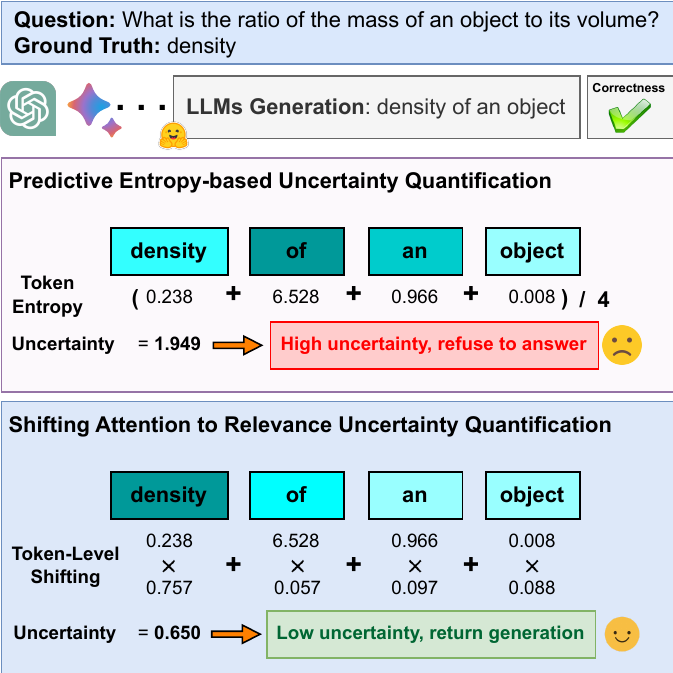}
    \caption{Irrelevant tokens (or sentences) may commit majority uncertainty in free-form generations, such as the token ``of'' committing extremely large uncertainty misleads the uncertainty quantification of LLMs. We term these observations as generative inequalities and tackle them by shifting attention to more relevant components.}
    \label{fig:represent_gene_inequal}
    \vspace{-4mm}
\end{figure}

Large Language Models (LLMs) have shown remarkable capabilities in multi-round conversation~\cite{long2023large,chen2023chatcot}, logical reasoning~\cite{creswell2022selectioninference,pan2023logiclm,duan2024gtbench}, and also disclose great potential in scientific discovery~\cite{Birhane2023ScienceIT}. For instance, ChatGPT, BARD, GPT-4, pre-trained on large-scale corpora and carefully aligned to human preferences~\cite{christiano2017deep, ouyang2022training}, profoundly shape the range of what AIs could do, and how they communicate with humans.

Despite the surprising progress, LLMs are proven to be vulnerable to widely known reliability issues~\cite{yao2024survey,sun2024trustllm,hong2024decoding}, such as hallucination~\cite{manakul2023selfcheckgpt} and factual errors~\cite{bian2023drop,karpinska2023large,gekhman2023trueteacher}. Uncertainty quantification (UQ) is one of the most popular approaches to answering when humans can trust the generations of LLMs, which is critical for Human-AI interaction applications (e.g., therapy and mental health~\cite{lin2023healthy,sharma2023human}) where humans need to densely communicate with LLMs. In these applications, the resulting behaviors will be largely affected by the generations from LLMs.

Unfortunately, UQ remains challenging due to various uncertainty sources (e.g., aleatoric uncertainty and epistemic uncertainty~\cite{kendall2017uncertainties}).
This challenge is particularly pronounced in the context of free-form LLMs, which are characterized by high complexity and an essentially limitless solution space—any output matching the semantic content of the true answer is considered correct. This makes UQ in LLMs markedly distinct from more traditional classification models or models with defined labels, where the solution space is constrained.

Prior works in this direction estimate uncertainty by prompting LLMs to answer confidence~\cite{lin2022teaching, Kadavath2022LanguageM} or designing logits- or entropy-based measurements~\cite{Malinin2021UncertaintyEI, malinin2020uncertainty, kuhn2023semantic}. The most recent work proposes \textit{Semantic Entropy} (\textit{SE})~\cite{kuhn2023semantic} where generations sharing the same meaning are gathered in a semantic cluster. Then the cluster-wise entropy is calculated as the uncertainty measurement.

Our motivation is derived from an intuitive fact: \textit{tokens are created unequally in presenting semantics}. Namely, some tokens (e.g., nouns, verbs) are more meaningful than other tokens (e.g., definite articles). For example, for a given question ``\textit{What is the ratio of the mass of an object to its volume?}'' and a model generation ``\textit{density of an object}'', ``density'' is the most relevant token in presenting semantics than the rest tokens. We term the former as \textit{relevant tokens} and the rest tokens as \textit{irrelevant tokens}. 
Prior works treat each token equally when estimating uncertainty, which is counter-intuitive (~\cref{fig:represent_gene_inequal}). Therefore, we ask:

\textit{Are relevant tokens more critical than irrelevant tokens in uncertainty quantification?}

To answer this question, we first investigate how token-level generative inequality affects uncertainty quantification in LLMs. Specifically, we first measure the \textit{relevance score} of each token by comparing the semantic change before and after removing this token from the sentence. A larger semantic change means more relevance for this token and vice versa. 
Then we quantify the \textit{uncertainty proportions}, i.e., the uncertainty committed by this token. At last, we analyze the correlation between relevance and uncertainty proportion. Our results reveal that large amounts of tokens containing very limited semantics are weighted equally or even heavily in UQ. Similar observations are also observed when generalizing to the sentence-level inequality by assessing relevant sentences and irrelevant sentences.

Based on these observations, we propose a simple attention-shifting method, by jointly examining the relevance of each component and reassigning its attention, from both the token level and the sentence level, termed as \textbf{S}hifting \textbf{A}ttention to \textbf{R}elevance (\textit{SAR}). \textit{SAR} is evaluated on multiple popular instruction-tuned LLMs (e.g., Vicuna~\cite{zheng2023judging}, LLaMA-2-chat~\cite{touvron2023llama}, WizardLM~\cite{xu2023wizardlm}), with model size up to 33B, and popular pre-trained LLMs (e.g., OPT~\cite{Zhang2022OPTOP}, LLaMA~\cite{Touvron2023LLaMAOA}) with model sizes up to 30b, over cross-domain free-form question-answering tasks, such as the conventional NLP domain (e.g.,  CoQA~\cite{reddy2019coqa}, TriviaQA~\cite{joshi2017triviaqa} and SciQ~\cite{Welbl2017CrowdsourcingMC}) and medical domain (e.g., MedQA~\cite{jin2020disease}, MedMCQA~\cite{pmlr-v174-pal22a}). Experimental results demonstrate \textit{SAR}'s superior performance. 
Our contributions can be summarized as the following:
\begin{itemize}
  \setlength\itemsep{-1pt} 

  \item We disclose that uncertainty quantification is significantly affected by token- and sentence-level generative inequality, i.e., irrelevant tokens or sentences might be over-valued when estimating uncertainty.
  \item We mitigate the two inequality biases by \textbf{S}hifting \textbf{A}ttention to \textbf{R}elevance (\textit{SAR}), which jointly examines the relevance of each token and sentence, and reassigns attention when estimating uncertainty.
  \item We conduct experiments over ``off-the-shelf'' instruction-tuned LLMs and popular pretrained LLMs, across various free-form question-answering tasks. Experimental results demonstrate that \textit{SAR} outperforms previous state-of-the-art by a large margin.
  
\end{itemize}

\section{Related Works}
\paragraph{Uncertainty Quantification in Conventional NLP Tasks.}
Uncertainty Quantification of machine translation (MT) has been studied for years to evaluate the performance of MT better. \cite{ott2018analyzing} access uncertainty by comparing multiple model outputs to multiple references with inter-sentence BLEU. \cite{glushkova2021uncertainty} measure uncertainty through techniques of Monte Carlo dropout~\cite{gal2016dropout} and deep ensembles~\cite{lakshminarayanan2017simple}.  
\cite{fomicheva2020unsupervised} use uncertainty quantification methods to improve probability estimates in neural networks.
\cite{lahlou2021deup} proposed Direct Epistemic Uncertainty Prediction, a model-agnostic framework, for estimating epistemic uncertainty in machine learning models.
For regression tasks, \cite{wang2022uncertainty} use uncertainty quantification to address both data uncertainty and model uncertainty, and
\cite{malinin2020regression} proposes a method for uncertainty quantification 
using Prior Networks to obtain interpretable measures of uncertainty at a low computational cost. 
For Natural Language Understanding tasks, \cite{talman2023uncertainty} use uncertainty quantification by applying Bayesian uncertainty modeling using Stochastic Weight Averaging-Gaussian.

\paragraph{Uncertainty Quantification in LLMs.}
Although uncertainty quantification has been thoroughly examined in models with distinct labels, such as classification models~\cite{ulmer2022exploring, vazhentsev2022uncertainty}, it is still under-explored for popular free-form LLMs, e.g., GPT~\cite{Radford2019LanguageMA}, OPT~\cite{Zhang2022OPTOP}, LLaMA~\cite{Touvron2023LLaMAOA}. These models present a unique challenge in uncertainty quantification as their solution domains are flexible and effectively infinite, i.e., any generation can be deemed correct as long as the semantics align consistently with the real answer.

\cite{xiao2022uncertainty} conducts large-scale empirical evaluations on how the configuration (e.g., model size, architecture, training loss) of LLMs affect uncertainty.
\cite{lin2022teaching, Kadavath2022LanguageM} propose to quantify uncertainty by directly prompting the language models to answer the uncertainty with respect to their generations. \cite{Manakul2023SelfCheckGPTZB} measures the faithfulness of generations by quantifying the consistency of generations, i.e., generations should be consistent if the model really captured the concept. \cite{Malinin2021UncertaintyEI} examines the uncertainty of free-form LLMs by calculating the accumulative predictive entropies over multiple generations. Recently, Semantic Entropy (\textit{SE})~\cite{kuhn2023semantic} is presented to tackle the ``semantic equivalence'' difficulty in uncertainty quantification. \textit{SE} gathers generations sharing the same semantics into clusters and performs cluster-wise predictive entropy as the uncertainty measurement.

We aim to design metrics from multiple generations to characterize the uncertainty of LLMs.
Our work focuses on the token- and sentence-level generative inequalities, which are not explored by prior works in uncertainty quantification.

\section{Generative Inequality in Uncertainty Quantification}

Tokens are created unequally in reflecting the meaning of the generation yet they are treated equally when estimating uncertainty. We term these inequalities as \textit{generative inequalities} and investigate how they affect uncertainty quantification.

\subsection{Preliminaries}\label{sec:generagive_inequ_preliminaries}
LLMs normally output generations in a free-form and auto-regressive manner, i.e., progressively predicting the probability distribution of the next token. We denote by $\vx$ the input (or the prompt) and $\vs$ the sentence consisting of $N$ tokens. Here, we take a sentence $\vs$ as a completion regarding prompt $\vx$. Then, for a given LLM, the probability of generating $z_i$ as the $i$-th token can be described as $p( z_i |\vs_{< i}, x) (1 \leq i \leq N)$, where $\vs_{< i}$ refers to the previously generated tokens $\{z_1, ..., z_{i - 1}\}$.

\noindent \textbf{Baseline.} We use the popular Predictive Entropy (\textit{PE}), described in~\cite{kadavath2022language}, as the baseline and investigate how it is affected by generative inequalities in this section. The Predictive Entropy (\textit{PE}) is defined as the entropy over the whole sentence $\vs$:
\begin{equation}\label{eq:PE}
    \textit{PE}(\vs, \vx) = - \log p(\vs|\vx) = \sum_{i}{-\log p(z_i|\vs_{< i}, \vx)}.
\end{equation}
It can be interpreted as the accumulation of the token-wise entropy.

\subsection{Token-Level Generative Inequality}\label{sec:token_level_inequ}
Generative inequality refers to an observation: tokens containing limited semantics are equally valued when estimating the uncertainty of a sentence, which is counter-intuitive. To outline this, we specify two quantities for each token: how much semantics the token contains, i.e., the \textit{relevance}, and how much uncertainty the token committed, i.e., the \textit{uncertainty proportion}.

For a given prompt $\vx$ and the sentence $\vs$ consisting of $N$ tokens, i.e., $\vs=\{z_1, z_2, ..., z_{\textit{N}}\}$:

\begin{figure}[t]
    \centering
    \includegraphics[width=\linewidth]{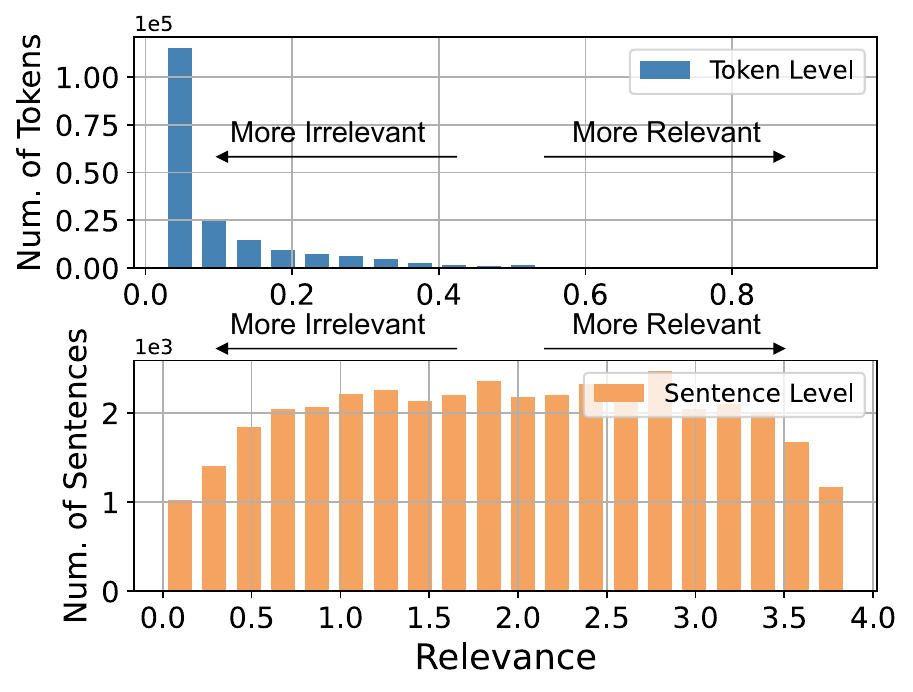}
    \caption{Distributions of relevance scores in both token-level and sentence-level situations. It is shown that irrelevant tokens and sentences take considerable proportions.}
    \label{fig:quantity_R_UP}
    \vspace{-1mm}
\end{figure}

\noindent \textbf{Relevance.} To measure how important $z_i$ is in reflecting the semantics of $\vs$, we compare the semantic change before and after removing this token: 
\begin{equation}\label{eq:token_relevance}
    \textit{R}_\textit{T}(z_i, \vs, \vx) = 1 - | g(\vx \cup \vs, \vx \cup \vs \setminus \{z_i\}) |,
\end{equation}
where $g(\cdot, \cdot)$, calculating the similarity between two sentences on a scale of 0 to 1, can be any semantic similarity measurement. In our experiments, we leverage the Cross-Encoder~\cite{reimers2019sentence}-RoBERTa-large~\cite{liu2019roberta} as this measurement since it is one of the most powerful sentence similarity evaluation models provided by the popular SentenceTransformers Library~\cite{reimers-2019-sentence-bert}. Generally, larger $\textit{R}_\textit{T}(z_i, \vs, \vx)$ means removing $z_i$ will lead to significant semantic changing, indicating that $z_i$ is more relevant.

\noindent \textbf{Uncertainty Proportion.} To measure the proportion of uncertainty committed by $z_i$, we simply derive the ratio from~\cref{eq:PE}:
\begin{equation}\label{eq:token_uncertainty_proportion}
    \textit{UP}_\textit{T}(z_i, \vs, \vx) = \frac{-\log p(z_i | \vs_{< i}, \vx)}{\textit{PE}(\vs, \vx)}. 
\end{equation}
Larger $\textit{UP}_\textit{T}(z_i, \vs, \vx)$ means $z_i$ commits more uncertainty when estimating the uncertainty of sentence $\vs$; vice versa.

\subsection{Sentence-Level Generative Inequality}\label{sec:sent_gene_inequal}
It has been widely shown that involving multiple sentences benefits estimating uncertainty~\cite{kadavath2022language}. For instance, \textit{PE} will usually be the arithmetic mean of multiple sentences in practice, i.e., $\frac{1}{K}\sum_{k}{\textit{PE}(\vs_k, \vx)} \, (1 \leq k \leq K)$ where $S=\{\vs_1, \vs_2, ..., \vs_{\textit{K}}\}$ consisting of $K$ sentences regarding $\vx$ and $ \vs_{k} \in S$ is the $k$-th sentence.
Following~\cref{sec:token_level_inequ}, for a given sentence $\vs_i$, we define the sentence-level relevance of $\vs_i$ as the probability-weighted semantic similarity with other sentences: 
\begin{equation}\label{eq:sentence_relevance}
    \textit{R}_{\textit{S}}(\vs_i, S, \vx) = \sum_{j=1,j \neq i}{g(\vs_i, \vs_j)p(\vs_j|\vx)},
\end{equation}
where $1 \leq i, j \leq K$ and $p(\vs_j | \vx)$ is the generative probability of $\vs_j$.
It is out of an intuitive assumption that sentences are more convincing if they are semantically consistent with other sentences.
Namely, a sentence that is semantically close to other sentences is considered more representative. Besides, the generative probability $p(\vs_j, \vx)$ provides more confidence to $\vs_j$ when measuring relevance, i.e., higher $p(\vs_j, \vx)$ makes $\vs_j$ more acceptable.

Similar to the token-level situation, the sentence-level uncertainty proportion of $\vs_i$ is defined as:
\begin{equation}\label{eq:sentence_uncertainty_proportion}
    \textit{UP}_\textit{S}(\vs_i, S, \vx) = \frac{\textit{PE}(\vs_i, \vx)}{\sum_{k}{\textit{PE}(\vs_k, \vx)}},
\end{equation}
where $1 \leq k \leq K$. It is the proportion of uncertainty committed by $\vs_i$,

\begin{figure}[t]
    \centering
    \includegraphics[width=0.95\linewidth]{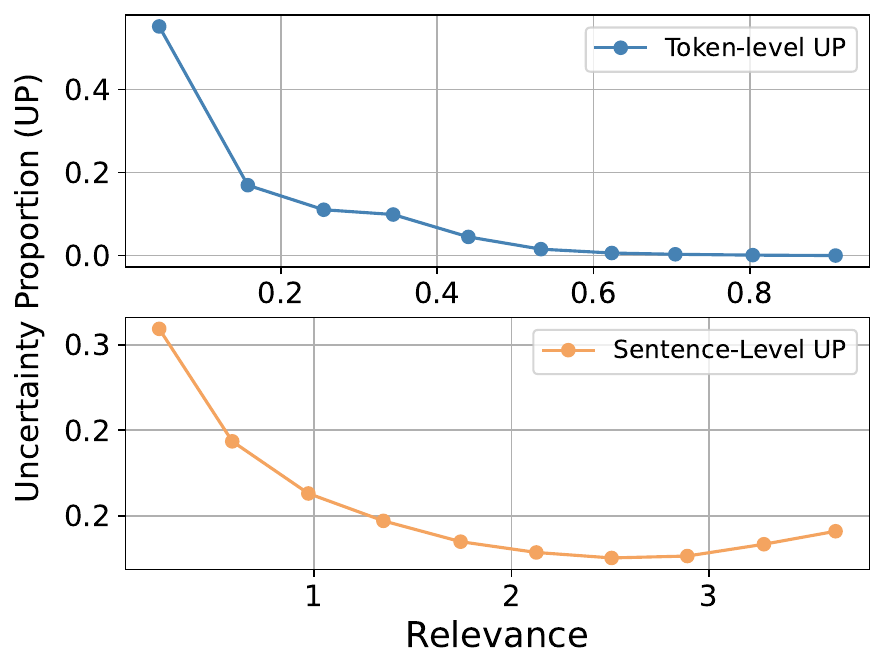}
    \caption{Correlations between relevance scores and uncertainty proportions in both token-level and sentence-level situations. Irrelevant tokens and sentences dominate the total volume of uncertainty quantification.}
    \label{fig:correclation_R_UP}
\end{figure}

\subsection{Analytical Insights}\label{sec:analytical_insights}
We leverage the defined \textit{relevance} and \textit{uncertainty proportion} to characterize the generative inequality observations in this section. 
We utilize CoQA as the dataset and OPT-13b as the model to be examined. For each prompt in CoQA, we generate 10 sentences, i.e., $K=10$ in~\cref{eq:sentence_relevance} and ~\cref{eq:sentence_uncertainty_proportion}. More details of generative hyper-parameters can be found in~\cref{sec:details_generative_inequa}. 

We first quantify the histograms of token-level relevance scores and sentence-level relevance scores. Results are summarized in~\cref{fig:quantity_R_UP}. For token-level relevance, it is clear that most of the tokens are irrelevant tokens, i.e., low relevance scores, indicating that linguistic redundancy exists widely. In terms of the sentence-level situation, although the distribution is smoother than the token-level situation, the irrelevant sentences still take a considerable amount over all the sentences.

We further investigate the correlations between relevance and uncertainty proportions, i.e., how much uncertainty is committed by tokens and sentences under various relevance scores. Specifically, we first group tokens and sentences into 10 bins with uniform relevance ranges and then average/sum the uncertainty proportions committed by tokens or sentences grouped in the same bin. Since irrelevant tokens take majority proportions over all the tokens, averaging the uncertainty proportions in each bin may hide the real effect of irrelevant tokens. Therefore, we report the sum of uncertainty proportions in each bin in the token-level situation.
Results are summarized in~\cref{fig:correclation_R_UP}.

It is clear that irrelevant tokens/sentences commit significantly more uncertainty than relevant sentences in both token-level and sentence-level situations. These observations demonstrate the existence of sentence inequalities and also the uncertainty quantification is highly affected by these inequalities.

\section{Shifting Attention to Relevance}

A natural hypothesis derived from~\cref{sec:analytical_insights} is that shifting the attention to those relevant components may benefit uncertainty quantification. In this section, we introduce the proposed Shifting Attention to Relevance (\textit{SAR}) in detail.

\subsection{Notations}
We reuse the notations defined in~\cref{sec:generagive_inequ_preliminaries} where we denote by $\vx$ the prompt and $S$ the $K$ sentences regarding $\vx$. There will be $N_j$ tokens for each sentence $\vs_j \in S \, (1 \leq j \leq K)$.

\subsection{Relevance Discovery and Shifting}
\textit{SAR} corrects generative inequalities by reviewing the relevance of each token and/or sentence and emphasizing uncertainty quantification attention to those more relevant components. Here we introduce token-level shifted measurement and sentence-level shifted measurements:

\noindent \textbf{Token-Level Shifting.} For a sentence $\vs_j$ regarding prompt $\vx$, $\vs_j = \{z_1, z_2, ..., z_{\textit{N}_j}\}$ contains $N_j$ tokens. We first calculate the normalized relevance score for each token $z_i \, ( 1 \leq i \leq N_j)$ based on~\cref{eq:token_relevance}, i.e., $\textit{R}_\textit{T}(z_i, \vs_j, \vx)$:
\begin{equation}\label{eq:token_relevance_norm}
    \tilde{\textit{R}}_\textit{T}(z_i, \vs_j, \vx) = \frac{\textit{R}_\textit{T}(z_i, \vs_j, \vx)}{\sum_{n}^{\textit{N}_j}{\textit{R}_\textit{T}(z_n, \vs_j, \vx)}}
\end{equation}
Then we enlarge the uncertainty proportions of relevant tokens by re-weighting token entropy according to their normalized relevance scores:
\begin{equation}\label{eq:shifted_token_entropy}
    \textit{E}_\textit{T}(z_i, \vs_j, \vx) = -\log p(z_i|\vs_{<i}, \vx) \tilde{R}_\textit{T}(z_i, \vs_j, \vx).
\end{equation}
The token-level shifted predictive entropy defined over $\vs_j$ can be formulated as:
\begin{equation}\label{eq:token_level_shifted_predictive_entropy}
    \textsc{token}{\textit{SAR}}(\vs_j, \vx) = \sum_{i}^{\textit{N}_j}{\textit{E}_\textit{T}(z_i, \vs_j, \vx)}.
\end{equation}
 The reason we normalize relevance score in~\cref{eq:token_relevance_norm} is two-fold: a) to make tokens comparable across sentences; b) to mitigate the bias posed by the length of sentence, such as the length normalization in Length-normalized Predictive Entropy (\textit{LN-PE})~\cite{malinin2020uncertainty}.
 In this way, the uncertainty proportions of tokens containing strong relevance will be enlarged when estimating uncertainty.

\begin{figure*}[t]
    \centering
    \includegraphics[width=\textwidth]{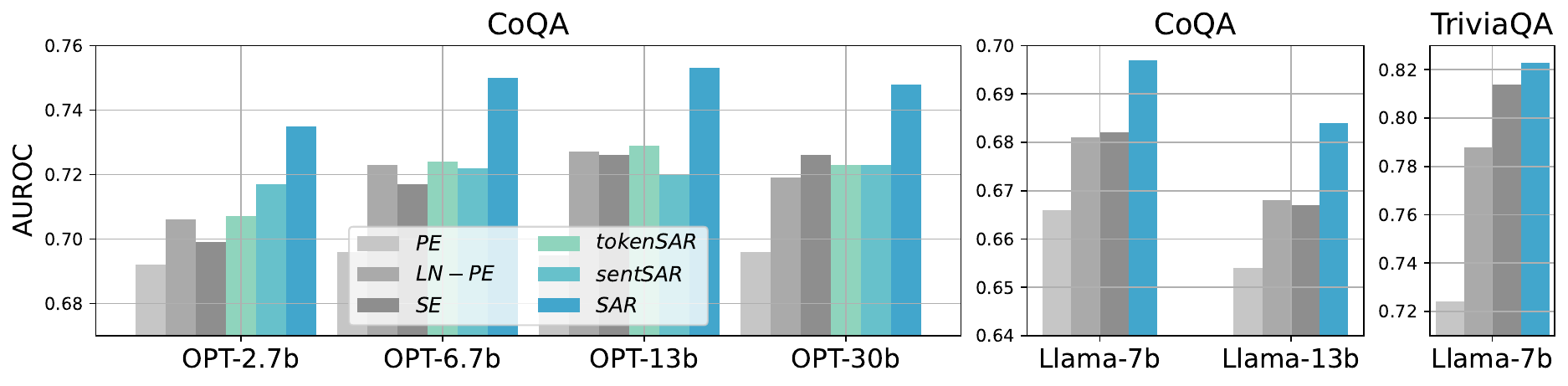}
    \caption{The AUROCs of \textsc{token}{\textit{SAR}}, \textsc{sent}{\textit{SAR}}, \textit{SAR}, and baseline methods, across various ``off-the-shelf'' LLMs and datasets (e.g., CoQA, and Trivia QA). Rouge-L with a threshold of 0.5 is used as the correctness metric. The proposed {\textit{SAR}} substantially outperforms existing methods across all the scenarios.}
    \label{fig:results_pretrained_llm}
\end{figure*}
\begin{table*}
    \centering
    \scriptsize
    \adjustbox{width=\textwidth}{
    
        \begin{tabular}{lcccc?ccc} \toprule
         \multirow{1}{*}{Models \& Datasets} & \multicolumn{1}{c}{\textit{LS}} & \multicolumn{1}{c}{\textit{PE}} & \multicolumn{1}{c}{\textit{LN-PE}} & \multicolumn{1}{c}{\textit{SE}} &  \multicolumn{1}{?c}{\textsc{token}{\textit{SAR}} ($\Delta$\textit{SE})} & \multicolumn{1}{c}{\textsc{sent}{\textit{SAR}}($\Delta$\textit{SE})} & 
         \multicolumn{1}{c}{\textit{SAR}($\Delta$\textit{SE})}\\
         \midrule
         \multicolumn{8}{l}{\textbf{Vicuna-13b} w./ 5 generations are generated for each question} \\
         \midrule
         Trivia QA & 0.560 & 0.690 & 0.624 & 0.630 & 0.692 (\color{ao(english)}{+6.2\%}) & \underline{0.745} (\color{ao(english)}{+11.5\%}) & \textbf{0.749} (\color{ao(english)}{+11.9\%}) \\
         SciQ & 0.589 & 0.708 & 0.668 & 0.675 & 0.706 (\color{ao(english)}{+3.1\%}) & \textbf{0.745} (\color{ao(english)}{7.0\%}) & \underline{0.741} (\color{ao(english)}{+6.6\%}) \\
         \midrule
         \multicolumn{8}{l}{\textbf{Vicuna-33b} w./ 5 generations are generated for each question} \\
         \midrule
         Trivia QA & 0.565 & 0.644 & 0.639 & 0.651 & 0.652 (\color{ao(english)}{+0.1\%}) & \textbf{0.715} (\color{ao(english)}{+6.4\%}) & \underline{0.710} (\color{ao(english)}{5.9\%}) \\
         SciQ & 0.584 & 0.665 & 0.668 & 0.674 & 0.665 (\color{amaranth}{-0.9\%}) & \textbf{0.717} (\color{ao(english)}{+4.3\%}) & \underline{0.710} (\color{ao(english)}{+3.6\%}) \\
         \midrule
         \multicolumn{8}{l}{\textbf{WizardLM-13b} w./ 5 generations are generated for each question} \\
         \midrule
         Trivia QA & 0.519 & 0.647 & 0.615 & 0.634 & 0.657 (\color{ao(english)}{+2.3\%}) & \underline{0.743} (\color{ao(english)}{+10.9\%}) & \textbf{0.744} (\color{ao(english)}{+11.0\%}) \\
         SciQ & 0.574 & 0.677 & 0.638 & 0.649 & 0.681 (\color{ao(english)}{+3.2\%}) & \textbf{0.719} (\color{ao(english)}{+7.0\%}) & \underline{0.707} (\color{ao(english)}{+5.8\%}) \\
         \midrule
         \multicolumn{8}{l}{\textbf{LLaMA-2-13b-chat} w./ 5 generations are generated for each question} \\
         \midrule
         Trivia QA & 0.504 & 0.647 & 0.615 & 0.622 & 0.654 (\color{ao(english)}{+3.2\%}) & \underline{0.698} (\color{ao(english)}{+7.6\%}) & \textbf{0.704} (\color{ao(english)}{+8.2\%}) \\
         SciQ & 0.578 & 0.718 & 0.688 & 0.692 & 0.718 (\color{ao(english)}{+2.6\%}) & \textbf{0.737} (\color{ao(english)}{+4.5\%}) & \underline{0.725} (\color{ao(english)}{+3.3\%}) \\
         \midrule
         \textbf{Average} & 0.555 & 0.675 & 0.644 & 0.653 & 0.678 (\color{ao(english)}{+2.5\%}) & \textbf{0.727} (\color{ao(english)}{+7.4\%}) & 0.724 (\color{ao(english)}{+7.1\%}) \\
         \bottomrule
    \end{tabular}
   
    }
    \caption{Uncertainty quantification AUROCs of \textsc{token}{\textit{SAR}}, \textsc{sent}{\textit{SAR}}, \textit{SAR}, and baseline methods, across various instruction-tuned open-source LLMs, over different datasets (e.g., SciQ, and Trivia QA). The threshold of Rouge-L is set to 0.5. Underline means the second best method.
    }
    \label{tab:instruction_tuned_model_results} 
    \vspace{-1mm}
\end{table*}

\noindent \textbf{Sentence-Level Shifting.} As mentioned in~\cref{sec:sent_gene_inequal}, sentences that have higher relevance scores, i.e., semantically consistent, are more convincing than others. Therefore, we simply reduce sentence uncertainty by enlarging its generative probability with a relevance-controlled quantity:
\begin{equation}\label{eq:sent_level_shifted_entropy}
    \begin{aligned}
        & \textit{E}_\textit{S}(\vs_j, S, \vx) = -\log ({p(\vs_j|\vx) + \frac{1}{t}\textit{R}_\textit{S}(\vs_j, S, \vx)})\\ &= -\log (p(\vs_j | \vx) + \underbrace{\frac{\sum_{k \neq j}{g(\vs_j, \vs_k){p(\vs_k | \vx)}}}{t}}_{\text{sentence relevance}}),
    \end{aligned}
\end{equation}
where $p(\vs_j | \vx) = \prod_{i}{p(z_i|\vs_{< i}, \vx)}$ is the generative probability of $\vs_j$ and $t$ is the temperature used to control the scale of shifting. 
Then, the sentence-level shifted predictive entropy over $K$ sentences can be formulated as:
\begin{equation}
    \textsc{sent}{\textit{SAR}}(S, \vx) = \frac{1}{K}\sum_{k}{\textit{E}_\textit{S}(\vs_k, S, \vx)}.
\end{equation}
Note that~\cref{eq:sent_level_shifted_entropy} shares a similar form with \textit{SE}~\cite{kuhn2023semantic}, i.e., reducing the uncertainty of semantically consistent sentences.
Differently, \textit{SE} achieves this with bi-directional entailment prediction and we achieve this with weighted relevance scores. With manual examination, we found that around 36.7\% of the entailment predictions are undesirable, over the long sentences that have more than 20 tokens on average (120 questions in total). Instead, our \textsc{sent}\textit{SAR} leverages the more ``soft'' sentence similarity to calculate the relevance score, which is more desirable for long and complex sentences.

\begin{figure*}
    \centering
    \includegraphics[width=\textwidth]{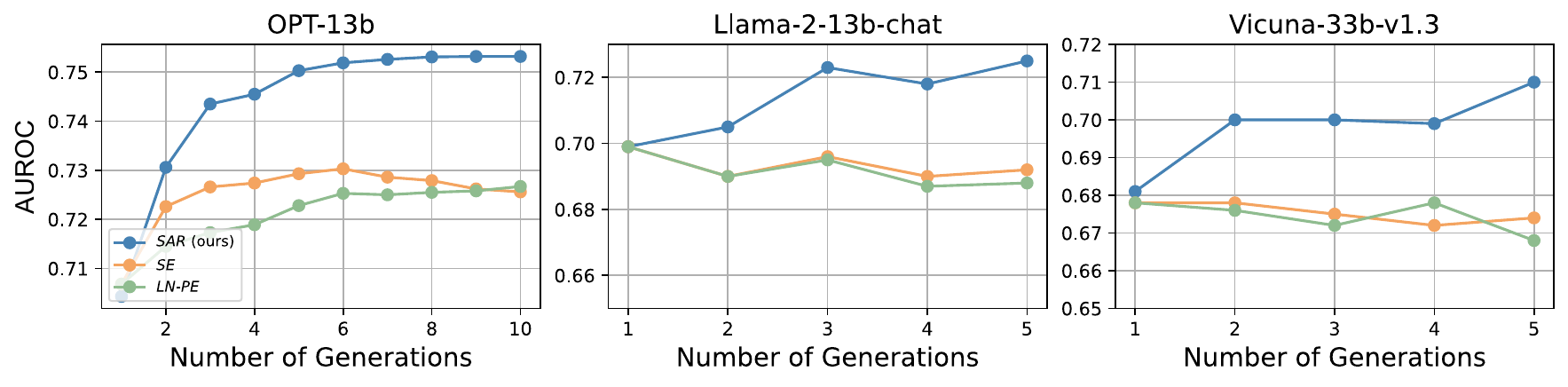}
    \caption{The performance of \textit{SAR} and baseline methods over various numbers of generations.}
    \label{fig:num_of_gens}
    \vspace{-1mm}
\end{figure*}
\begin{figure*}
    \centering
    \includegraphics[width=\textwidth]{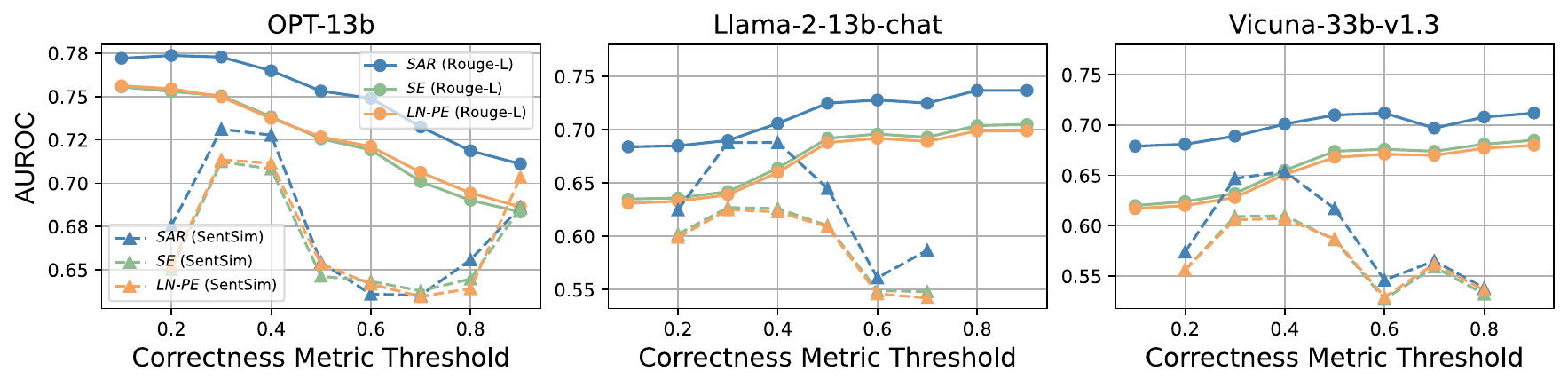}
    \caption{The performance of \textit{SAR} over various Rouge-L and Sentence Similarity thresholds.}
    \label{fig:rouge_L_sensitivity}
    \vspace{-1mm}
\end{figure*}
\subsection{Overall Measurement}
Token-level shifting and sentence-level shifting are conceptually different as they emphasize different perspectives. However, they are orthogonal and can be naturally combined to shift attention from both token-level and sentence-level, resulting in more effective uncertainty quantification. To achieve that, we simply replace the generative probabilities in ~\cref{eq:sent_level_shifted_entropy}, i.e., $p(\vs_i | \vx)$ and $p(\vs_j | \vx)$, with the token-shifted probability derived from \cref{eq:token_level_shifted_predictive_entropy},
i.e. $p^\prime(\vs_i | \vx) = e^{-\textsc{token}{\textit{SAR}}(\vs_i, \vx)}$ and $p^\prime(\vs_j | \vx) = e^{-\textsc{token}{\textit{SAR}}(\vs_j, \vx)}$:
\begin{equation}\label{eq:sar_shifted_entropy}
\small
        \textit{E}_\textit{T,S}(\vs_j, S, \vx) = -\log (p^\prime(\vs_i | \vx) + \frac{\sum_{k \neq j}{g(\vs_j, \vs_k)p^\prime(\vs_j | \vx)}}{t}).
\end{equation}
Then the token- and sentence-level shifted predictive entropy over $K$ sentences can be defined as $ \textit{SAR} = \frac{1}{K}\sum_{k}\textit{E}_\textit{T,S}(\vs_k, S, \vx)$.

We denote \textsc{token}{\textit{SAR}}, \textsc{sent}{\textit{SAR}}, and \textit{SAR} as the token-shifted predictive entropy, sentence-shifted predictive entropy, and both token- and sentence-shifted predictive entropy respectively, in the rest of this paper.

\section{Empirical Evaluations}

\subsection{Experimental Settings}

\noindent \textbf{Baselines.} We consider 4 baseline methods in our experiments, including Lexical Similarity~\cite{lin2022towards}, Semantic Entropy (\textit{SE})~\cite{kuhn2023semantic}, Predictive Entropy (\textit{PE})~\cite{kadavath2022language}, and Length-normalized Predictive Entropy (\textit{LN-PE})~\cite{malinin2020uncertainty}. Lexical Similarity considers the similarities among multiple sentences. \textit{SE} introduces the ``semantic equivalence'' difficulty in the uncertainty quantification of free-form LLMs and tackles this issue by gathering sentences containing the same meaning into clusters and calculating cluster-wise entropy. \textit{LN-PE} is the length normalized \textit{PE}, i.e., divided by sentence length $N$: $\textit{LN-PE} (\vs, \vx) = \frac{1}{N} \textit{PE} (\vs, \vx)$. 

\noindent \textbf{Models.} We conduct experiments over popular ``off-the-shelf'' LLMs, including instruction-tuned LLMs (e.g., Vicuna~\cite{zheng2023judging}, LLaMA-2-chat~\cite{touvron2023llama}, WizardLM~\cite{xu2023wizardlm}) and pre-trained LLMs (e.g., OPT~\cite{Zhang2022OPTOP} and LLaMA~\cite{Touvron2023LLaMAOA}), with model size up to 33B. 
We leverage greedy search for the most likely generations which are used to evaluate correctness, and multinominal sampling for reference generations which are used to estimate uncertainty. 
More details of generative hyper-parameters can be found in~\ref{sec:details_generative_inequa}

\noindent \textbf{Datasets.} We consider 5 free-form question-answering datasets: CoQA~\cite{reddy2019coqa}, Trivia QA~\cite{joshi2017triviaqa}, SciQ~\cite{Welbl2017CrowdsourcingMC}, MedQA~\cite{jin2021disease} and MedMCQA~\cite{pmlr-v174-pal22a}. More details of the used datasets and the splittings can be found in~\ref{sec:dataset}.

\noindent \textbf{Correctness Metrics.} We adopt Rouge-L~\cite{Lin2004ROUGEAP} and sentence similarity as the correctness metrics when evaluating the correctness of LLMs' generations. We set the threshold of Rouge-L and sentence similarity as 0.5, i.e., generations having above 0.5 semantic similarities/Rouge-L scores with the ground truth are correct. Sentence similarity is measured by DistillRoBERTa~\cite{Sanh2019DistilBERTAD} in SentenceTransformers~\cite{reimers-2019-sentence-bert}.  The sensitivity of \textit{SAR} to these thresholds will be studied in~\cref{sec:ablation_study}.

\noindent \textbf{Evaluation Metric.} Following prior work~\cite{kuhn2023semantic}, we evaluate uncertainty quantification by predicting the correctness of the model's generations regarding a given question. The area under the receiver operator characteristic curve (AUROC) indicates the probability that a random correct generation has a lower uncertainty than a random incorrect generation, predicted by uncertainty quantification methods. 
AUROC equals 0.5 meaning the assigned uncertainty is no better than random guessing, i.e., they can not differentiate between correct and incorrect generations. AUROC equals 1 meaning all the correct generations are assigned lower uncertainty than all incorrect generations.

\noindent \textbf{Hyperparameters.} For OPT-2.7b/6.7b/13b, we generate 10 generations for each question, i.e. $K$=10. For other models, we generate 5 generations. The temperature $t$ is set to 0.001. 
The generative settings can be can be found in~\cref{sec:details_generative_inequa}. All the experiments are conducted on a server with one Intel(R) Xeon(R) Platinum 8358 CPU and two NVIDIA A100 GPUs.

\subsection{UQ for pre-trained LLMs}
We compare \textit{SAR}, \textsc{token}{\textit{SAR}}, and \textsc{sent}{\textit{SAR}} with state-of-the-art methods. Results are summarized in~\cref{fig:results_pretrained_llm}.
Generally, our methods significantly outperform prior methods in most of the settings. For instance, \textit{SAR} outperforms other methods by at most 3.6\% AUROC over the CoQA dataset, measured by Rouge-L 0.5. The results of setting Rouge-L to 0.3, which is the same as in~\cite{kuhn2023semantic}, can be found in~\cref{sec:main result rl 0.3}.

\begin{table}[t]
    \centering
    \scriptsize
    \adjustbox{width=\linewidth}{
    \begin{tabular}{cc|ccccc}
        \toprule
        & & \multicolumn{4}{c}{\textit{SAR} w. sentence similarity } \\
         OPT Size & \textit{SE} &  RoBERTa & MiniLM & MPNet & OPT-13b \\
          \midrule
        2.7b & 0.699 & \textbf{0.735} & 0.723 & 0.723 & 0.716\\
         6.7b & 0.717 &  \textbf{0.750} &0.740&0.739&0.731 \\
         13b & 0.725 & \textbf{0.753}&0.741&0.740&0.733 \\
         30b & 0.726 & \textbf{0.748}&0.738&0.739&0.734 \\
          \bottomrule
    \end{tabular}
    }
    \caption{Sensitivity of \textit{SAR} to sentence similarity measurements. We consider popular models from SentenceTransformers (\cref{appendix:sentence_similarity}) and also the target LLMs as the sentence similarity measurement.}
    \label{tab:sentence_similarity}
    \vspace{-1mm}
\end{table}

Also, the synergy of \textsc{token}{\textit{SAR}} and \textsc{sent}{\textit{SAR}} achieves remarkable improvements. For instance, \textsc{token}{\textit{SAR}} and \textsc{sent}{\textit{SAR}} achieve 0.723 AUROC in the OPT-30b-CoQA setting yet combining them results in 0.748 AUROC. It indicates that \textsc{token}{\textit{SAR}} and \textsc{sent}{\textit{SAR}} are compatible and can be incorporated effectively. 

\subsection{UQ for Instruction-Tuned LLMs}
We estimate the uncertainty of powerful instruction-tuned LLMs, including Vicuna-13b/33b, LLaMA-2-chat-13b, and WizardLM-13b. All these models are obtained from Huggingface, without any further modifications.
Results are summarized in~\cref{tab:instruction_tuned_model_results}. It is shown that \textit{SAR} consistently beat baseline methods in most situations. 
For example, \textit{SAR} outperforms \textit{SE} by 7.1\% AUROC on average, evaluated by Rouge-L 0.5. 

\subsection{Ablation Studies}\label{sec:ablation_study}

\noindent \textbf{Number of Generations.} The effects of the number of generations are summarized in~\cref{fig:num_of_gens}. It is shown that our \textit{SAR} is generation-efficient, i.e., it achieves 0.750 AUROC with only 5 generations and it can be consistently boosted with more generations, while other methods may even drop slightly when more generations are provided.

\noindent \textbf{Sensitivity to Sentence Similarity.} We investigate the sensitivity of SAR to sentence similarity measurements. As shown in~\cref{tab:sentence_similarity}, general-purpose sentence similarity models are desirable and more effective than the target LLMs (last column of~\cref{tab:sentence_similarity}). This is because LLMs are not specifically designed for sentence similarity.

\begin{table}[t]
    \centering
    \scriptsize
    \adjustbox{width=\linewidth}{
    \begin{tabular}{lccc}
        \toprule
         Method & \# Generation & Time (s) & \makecell[c]{avg. AUROC \\ SciQ/Trivia QA}  \\
         \midrule
         \textit{PE} & 5 & 5.28 & 0.692/0.657 \\
         \textit{LN}-\textit{PE} & 5 & 5.28 & 0.666/0.623 \\
         \textit{SE} & 5 & 6.78 & 0.673/0.634 \\
         \midrule
         \textsc{sent}{\textit{SAR}} & 2 & \textbf{2.64} & \textbf{0.708}/\textbf{0.685} \\
         \bottomrule
    \end{tabular}
    }
    \caption{Efficiency comparisons between 2-generations \textit{SAR} and 5-generations baseline methods.}\label{tab: two_generations_sentsar}
\end{table}
\begin{table}
    \centering
    \scriptsize
    \adjustbox{width=\linewidth}{
    \begin{tabular}{cc|ccc}
        \toprule
         Model  & Dataset &  \textit{LN-PE} & \textit{SE} & \textit{SAR} \\
          \midrule
        \multirow{2}{*}{Vicuna-13b} & MedQA & 0.572 & \textbf{0.599} & 0.598 \\
         & MedMCQA & 0.649 & 0.685 & \textbf{0.717} \\
        \midrule
        \multirow{2}{*}{LLaMA-2-13b-chat} & MedQA  & 0.562 & 0.609 &\textbf{ 0.616} \\
        & MedMCQA & 0.647 & 0.655 & \textbf{0.702} \\
        \midrule
        WizardLM-13b & MedQA & 0.609 & 0.620 & \textbf{0.635}\\
          \bottomrule
    \end{tabular}
    }
    \caption{The performance of \textit{SAR} and baseline methods over medical Q\&A datasets. Our method achieves better performances for most settings.}
    \label{tab:medical_results}
    \vspace{-1mm}
\end{table}

\noindent \textbf{Sensitivity to Correctness Metrics.} Empirical results are presented in~\cref{fig:num_of_gens}. Higher thresholds mean the correctness standards are more harsh. It is shown that the performances of uncertainty quantization will be affected as the metrics are getting harsh. However, our methods significantly outperform baseline methods in most cases.

\noindent \textbf{Efficiency Comparison.} In~\cref{sec:appendix_computational_cost}, we provide a detailed computational cost analysis, regarding the time consumed by each operation. We provide the results of 2-generations \textsc{sent}\textit{SAR} with 5-generations baseline methods in~\cref{tab: two_generations_sentsar} over instruction-tuned LLMs. Our method surpasses the baseline methods while consuming less than half the time, demonstrating its greater generation efficiency.

\subsection{UQ in Medical Domain}
We evaluate \textit{SAR} over the AI for science scenarios, such as medical domains. As shown in~\cref{tab:medical_results}, we perform experiments over MedQA~\cite{jin2020disease} and MedMCQA~\cite{pmlr-v174-pal22a} datasets and our methods achieve better performance for most of the settings. This indicates the potential impacts of our methods on the real world.

\vspace{-1mm}
\section{Conclusion}
In this paper, we disclose the generative inequality observation in uncertainty quantification: tokens and generations are created unequally in reflecting semantics yet they are treated equally when estimating uncertainty, which is counter-intuitive. We propose to tackle these inequalities by Shifting Attention to Relevance (\textit{SAR}) from both token-level (\textsc{token}{\textit{SAR}}) and sentence-level (\textsc{sent}{\textit{SAR}}). Experiments over ``off-the-shelf'' LLMs demonstrate the superior performances of \textit{SAR}. 

\section{Ethical Considerations}
Our proposed method has the potential to impact the credibility and reliability of LLMs, particularly in the context of reducing misinformation.
LLMs have the potential to generate highly plausible but false information. Uncertainty quantification techniques can help distinguish between accurate and misleading outputs. Success in adequately addressing this issue can contribute to the prevention
spread of misinformation and its potential societal consequences

\section{Limitations}
Our method will introduce sentence similarity calculations and comparisons. We tackle this issue by leveraging a small backbone in our implementation but it still might bring additional latency in practice. In addition, our methods require access to token logits. Although token logits are
widely supported by commercial LLM providers, this still might restrict the potential application of our methods in black-box scenarios.

\section*{Acknowledgement}
This work was performed under the auspices of the U.S. Department of Energy by Lawrence Livermore National Laboratory under Contract DE-AC52-07NA27344 and was supported by the LLNL-LDRD Program under Project No. 23-ERD-030 (LLNL-CONF-851171). This work was partially supported by the National Science Foundation under Grant No.2319242.

\bibliography{custom}

\clearpage
\newpage

\appendix
\section*{Appendix}

\section{Details of LLMs Generation}\label{sec:details_generative_inequa}

\noindent \textbf{OPT models.} We will generate 1 most likely generation with the greedy search for all the OPT models. This generation will be used to evaluate the correctness. 
For OPT-2.7b/6.7b/13b, we will generate 10 sentences for each question with multinomial sampling for uncertainty quantification. For OPT-30b, we will generate 5 sentences. The temperature of generation is fixed at 0.5 for all models. For OPT-2.6b/6.7b/13b, the max length of each generation is set to 256 tokens for the CoQA dataset and SciQ dataset and is set to 128 tokens for the Trivia QA dataset. For OPT-30b, the max length of each generation is set to 128 tokens for all the datasets.

\noindent \textbf{LLaMA/Vicuna/WizardLM.} We will generate 1 most likely generation with the greedy search and 5 sentences with multinomial sampling for all these models. The max length of each generation is set to 128 tokens. The temperature of generation is set to 0.5.

\section{Datasets}\label{sec:dataset}
\textbf{CoQA}~\cite{reddy2019coqa} is a large-scale conversational QA task, with more than 127,000 questions. Each question is equipped with a passage to provide contextual information.
\textbf{Trivia~QA}~\cite{joshi2017triviaqa} is a high-quality reading comprehension dataset that contains over 650k question-answer pairs. These questions are obtained from trivia enthusiasts and answers from Wikipedia. 
\textbf{SciQ}~\cite{Welbl2017CrowdsourcingMC} dataset is a science-related QA dataset aimed at developing models' capabilities of understanding complex scientific texts. It consists of approximately 13,679 crowdsourced science questions. 
\textbf{MedQA}~\cite{jin2020disease} is a free-form multiple-choice OpenQA dataset for solving medical problems, collected from the professional medical board exams. \textbf{MedMCQA}~\cite{pmlr-v174-pal22a} is a large-scale, Multiple-Choice Question Answering (MCQA) dataset designed to address real-world medical entrance exam questions.

Following~\cite{kuhn2023semantic}, we randomly select around 8,000 questions from the training
split of Trivia QA as the questions to be examined. For instruction-tuned experiments, we use 2,000 questions of Trivia QA. We utilize
the full validation set (1,000 questions) of SciQ and the development split (7,983 questions) of CoQA.

\section{Additional Experimental Analysis}\label{sec:ablation}

\subsection{Effects of \textit{SAR} Temperature $t$}\label{sec:effect_t}
The hyperparameter $t$ introduced in~\cref{eq:sent_level_shifted_entropy} is used to control the scale of sentence shifting. The effects of $t$ is provided in~\cref{tab:effect_of_temperature}. It is shown that $t$ marginally affects the performance of \textit{SAR}.

\begin{table}[h]
    \centering
    
    \adjustbox{width=\linewidth}{
    \begin{tabular}{ccccc}
        \toprule
        \multirow{2}{*}{$t$} & \multicolumn{2}{c}{OPT-13b} & \multicolumn{2}{c}{LLaMA-7b} \\
        \cmidrule(lr){2-3}
        \cmidrule(lr){4-5}
        & CoQA & SciQ & CoQA & TriviaQA \\
        \midrule
        \multicolumn{1}{l}{$1\times10^{-3}$} & 0.753/0.720 & 0.737/0.784 & 0.697/0.658 & 0.823/0.815 \\
        \multicolumn{1}{l}{$1\times10^{0}$} & 0.752/0.719 & 0.739/0.786 & 0.695/0.656 & 0.822/0.816 \\
        \multicolumn{1}{l}{$1\times10^{1}$} & 0.743/0.714 & 0.729/0.786 & 0.686/0.658 & 0.813/0.812 \\
        \bottomrule
    \end{tabular}
    }
    \caption{Effects of temperature $t$ in~\cref{eq:sent_level_shifted_entropy}. Results are evaluated by Rouge-L with 0.5 as the threshold. Results are obtained from \textit{SAR}/\textsc{token}{\textit{SAR}}.}
    \label{tab:effect_of_temperature}
\end{table}

\subsection{Generation Efficiency}
\begin{figure}[h]
    \centering
    \includegraphics[width=\linewidth]{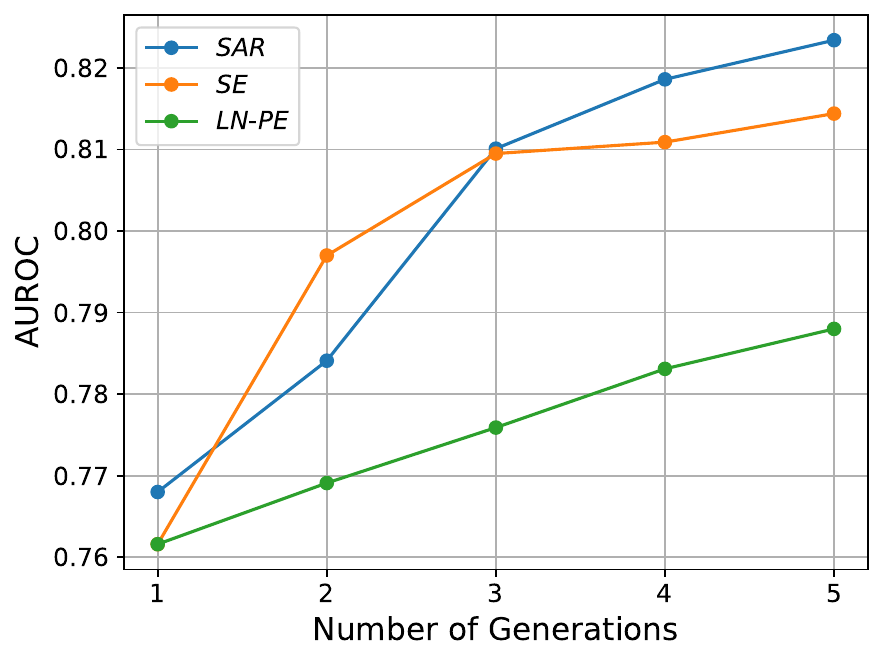}
    \caption{The performance of \textit{SAR} over various numbers of generations. Results are obtained from the LLaMA-7b model over the Trivia QA dataset.}
    \label{fig:llama7b_trivia_qa_num_generation}
\end{figure}

The generation-efficiency of \textit{SAR} on LLaMA-7b-Trivia QA setting is presented in~\cref{fig:llama7b_trivia_qa_num_generation}.

\subsection{Sensitivity to Sentence Length.}
\begin{figure}[h]
    \centering
    \includegraphics[width=0.5\textwidth]{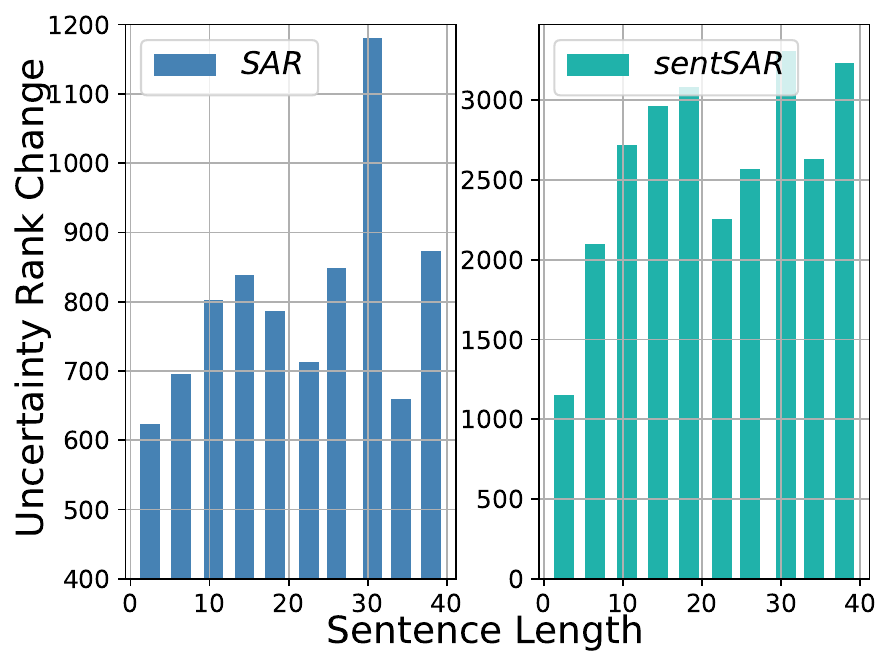}
    \caption{Demographic analysis of sentence length. Uncertainty Rank Change between (\textbf{Left}) \textit{SAR} and \textit{LN-PE}, and between (\textbf{Right}) \textsc{sent}{\textit{SAR}} and \textit{LN-PE}. It is shown that \textit{SAR} and \textsc{sent}{\textit{SAR}} are more tend to affect medium- or long-length sentences.}
    \label{fig:sentence_length_demographic_analysis}
\end{figure}

To study how the \textit{SAR} is affected by sentence length, we quantify the uncertainty rank change for each sentence, caused by \textit{SAR} and \textsc{sent}{ß\textit{SAR}}. Assume a sentence has a rank of $i$ among all the sentences, evaluated by LN-PE and has a rank of $j$ evaluated by \textit{SAR}, then the uncertainty rank change is $|i - j|$. The correlations between average uncertainty rank change and sentence length are presented in~\cref{fig:sentence_length_demographic_analysis}.
It is shown that our methods tend to conclude medium- and long-length sentences.

\subsection{Different Correctness Metric Threshold}\label{sec:main result rl 0.3}

We report the results of Rouge-L (0.3) (same as~\cite{kuhn2023semantic} in~\cref{tab: main result rl 0.3}.

\begin{table*}[!htp]\centering

\scriptsize

\adjustbox{width=\textwidth}{
\begin{tabular}{llcccc|cccc}\toprule
Dataset &Model &\textit{LS} &\textit{PE} &\textit{LN-PE} &\textit{SE} &\textsc{token}\textit{SAR} &\textsc{sent}\textit{SAR} &\textit{SAR} \\\midrule
\multirow{6}{*}{CoQA} &OPT-2.7b &0.573 &0.666 &0.719 &0.712 &0.719 &0.689 &\textbf{0.742} \\
&OPT-6.7b &0.588 &0.671 &0.745 &0.741 &0.746 &0.696 &\textbf{0.768} \\
&OPT-13b &0.588 &0.666 &0.750 &0.751 &0.752 &0.690 &\textbf{0.773} \\
&OPT-30b &0.550 &0.671 &0.742 &0.751 &0.746 &0.698 &\textbf{0.767} \\
\cmidrule{2-9}
&LLaMA-7b &0.511 &0.646 &0.673 &0.672 &0.672 &0.635 &\textbf{0.686} \\
&LLaMA-13b &0.522 &0.617 &0.653 &0.652 &0.653 &0.610 &\textbf{0.665} \\
\midrule
\multirow{2}{*}{Trivia QA} &LLaMA-7b &0.533 &0.713 &0.783 &0.814 &0.793 &0.800 &\textbf{0.818} \\
&LLaMA-13b &0.655 &0.492 &0.627 &\textbf{0.758} &0.635 &0.749 &0.716 \\
\midrule
\multicolumn{2}{c}{\textbf{Average}} &0.565 &0.643 &0.712 &0.731 &0.715 &0.696 &\textbf{0.742} \\
\bottomrule
\end{tabular}
}
\caption{Uncertainty estimation AUROCs of \textsc{token}{\textit{SAR}}, \textsc{sent}{\textit{SAR}}, \textit{SAR}, and baseline methods, across various ``off-the-shelf'' LLMs and datasets (e.g., CoQA, and Trivia QA). Rouge-L with a threshold of 0.3 is used as the correctness metric.}\label{tab: main result rl 0.3}
\end{table*}

\subsection{Computational Costs Analysis}\label{sec:appendix_computational_cost}
\textit{SAR} is more generation-efficient. It surpasses baseline methods under significantly smaller computational constraints. We have quantified the time consumed for each step in the overall uncertainty quantification pipeline. This includes sequence generation, computing logits, semantic clustering for SE, and sentence similarity for \textit{SAR}. We exclude the time taken for aggregating logits/scores as it is negligible (less than 0.001 seconds for all methods). The average time consumed per question, based on an evaluation of 1000 questions from the Vicuna-13b + SciQ dataset, is provided. These measurements were taken using an AMD EPYC 7302 16-Core CPU and a 1xA40 GPU server. Results are summarized in~\cref{tab: computational costs}.

\begin{table*}[!htp]\centering

\adjustbox{width=\textwidth}{
\begin{tabular}{lcccccc}\toprule
Method &Num. of Generations &Generation &Logits Computing &Semantic Clustering &Sentence Similarity &Sum \\\midrule
\textit{PE} &5 &4.09s &1.19s &0s &0s &5.28s \\
\textit{LN-PE} &5 &4.09s &1.19s &0s &0s &5.28s \\
\textit{SE} &5 &4.09s &1.19s &1.5s &0s &6.78s \\
\midrule
\textsc{sent}\textit{SAR} &5 &4.09s &1.19s &0s &2.58s &7.86s \\
\textsc{sent}\textit{SAR} &2 &1.64s &0.48s &0s &0.52s &\textbf{2.64s} \\
\bottomrule
\end{tabular}
}
\caption{Computational costs of \textit{SAR} and baseline methods. We report both \textsc{sent}\textit{SAR} with 5 and 2 generations.}\label{tab: computational costs}
\end{table*}

\section{Sentence Similarity Measurement}\label{appendix:sentence_similarity}
The following is the sentence similarity measurement models we leveraged in~\cref{tab:sentence_similarity}:
\begin{itemize}
    \item RoBERTa: cross-encoder/stsb-roberta-large
    \item MiniLM: sentence-transformers/all-MiniLM-L6-v2
    \item MPNet: sentence-transformers/all-mpnet-base-v2
\end{itemize}

\end{document}